\documentclass[10pt,twocolumn,letterpaper]{article}

\usepackage{cvpr}      

\usepackage[tracking=true]{microtype}
\usepackage{xspace}
\usepackage{microtype}
\usepackage{xspace}
\usepackage{fontawesome5}
\usepackage{xcolor}
\definecolor{projectpink}{HTML}{E75480}
\newcommand{\genvla}{{\fontfamily{ptm}\mdseries\scshape\selectfont Genvla}\xspace}

\usepackage{pifont}
\usepackage{tabularx}
\usepackage{multirow}
\usepackage[table]{xcolor}
\newcommand{\stitle}[1]{\vspace*{0.4em}\noindent{\bf #1\/}}
\usepackage{wrapfig}
\usepackage{cuted} 

\usepackage{currfile} 

\usepackage{caption} 
\definecolor{cvprblue}{rgb}{0.21,0.49,0.74}
\usepackage[pagebackref,breaklinks,colorlinks,allcolors=cvprblue]{hyperref}

\def\confName{CVPR}
\def\confYear{2027}

\title{\textbf{\genvla}: Execution-Conditioned Parameter Generation for Vision-Language-Action Policies}

\author{
Zhaofeng Hu$^{1,\ddagger}$ \quad
Leyao Wang$^{2,\ddagger}$ \quad
Minghui Wu$^{3,4}$ \quad
Tianyu Fu$^{3}$ \\
Chenxu Zhao$^{3,*}$ \quad
Qinbo Zhang$^{1}$ \quad
Haoyu Chen$^{2}$ \quad
Kai Wang$^{5}$ \\
Sifan Zhou$^{6}$ \quad
Xiaobo Wang$^{2,*}$ \quad
Ci-Jyun Liang$^{1}$ \quad
Shuo Wang$^{3,\dagger}$ \\[4pt]
$^{1}$Stony Brook University \quad
$^{2}$Shenzhen University of Advanced Technology \\
$^{3}$Mininglamp Technology \quad
$^{4}$Peking University \\
$^{5}$National University of Singapore \quad
$^{6}$Carnegie Mellon University \\[3pt]
{\small
$^{\ddagger}$Equal contribution \quad
$^{*}$Corresponding authors \quad
$^{\dagger}$Project lead} \\[4pt]
{\small
\href{https://windyhu001.github.io/GenVLA/}
{\textcolor{projectpink}{\faGlobe\ \textbf{Project Website}}}}
}
\begin{document}

\typeout{FONT FAMILY: \rmdefault}
{\normalfont\typeout{BODY FONT: \fontname\font}}
{\bfseries\typeout{BOLD FONT: \fontname\font}}

\maketitle
\begin{abstract}
Pretrained Vision-Language-Action (VLA) models provide a powerful foundation for robot manipulation, yet adapting them to heterogeneous control demands remains challenging. Such heterogeneity arises not only across tasks but also within individual executions, where manipulation transitions through approach, grasping, transport, and placement, each requiring distinct control behaviors. Conventional low-rank adaptation (LoRA) relies on static parameter updates, while temporally adaptive variants remain constrained by predefined conditioning structures. We introduce \genvla, an execution-conditioned parameter generation framework that dynamically synthesizes low-rank updates for a VLA's vision-language backbone and action expert from current visual-language features and short-term execution history. By generating distinct updates throughout
closed-loop execution, \genvla constructs an
expressive parameter family from individually
low-rank adaptations. Importantly, these generated updates exhibit structured variations associated with evolving manipulation stages, without requiring explicit phase supervision or online optimization. Experiments on LIBERO, RoboTwin~2.0, and real-world manipulation demonstrate strong performance, achieving 93.0\% average success rate on LIBERO with rank-4 updates. Further ablations and parameter-space analyses validate the benefits of dynamic generation and reveal execution-dependent adaptation patterns.  Code will be available soon. 
\end{abstract}    

\section{Introduction}

Pretrained Vision-Language-Action (VLA) models have become a powerful foundation for robot manipulation~\citep{brohan2022rt1,zitkovich2023rt2,kim2024openvla,black2025pi0,black2025pi05}, while parameter-efficient fine-tuning (PEFT)~\citep{houlsby2019parameter}, particularly low-rank adaptation (LoRA)~\citep{hu2021lora}, enables their efficient adaptation to downstream tasks. Yet robot manipulation is heterogeneous, both across tasks and within individual executions. Tasks such as grasping, pushing, transporting, and precise placement demand distinct behaviors. Even under the same instruction, a robot may transition through approach, alignment, contact, grasping, and placement, each requiring different perceptual cues and control precision. This raises a central challenge for VLA adaptation:

\emph{How can low-rank adaptation accommodate heterogeneous control demands both across tasks and within individual executions?}

Static LoRA methods apply the same learned parameter update across different control contexts, forcing heterogeneous control demands into a single shared adaptation. This can lead to compromised parameter updates that limit adaptation flexibility and expressivity, a phenomenon related to the \emph{shared parameter bottleneck}~\citep{jacob2026copra,wu2026hy}. Other VLA approaches address heterogeneity through task or embodiment conditioning, expert routing, and modular adaptation~\citep{wang2024hpt,zheng2025uniact,zheng2025xvla,du2025himoevla,shen2025adamoe,zhang2026corevla,luo2026coral}. Recent work has explored adaptation within a trajectory. PhaseLoRA~\citep{guo2026phaselora} focuses on within-trajectory adaptation by modulating shared LoRA components through two control descriptors. While this enables temporal adaptation, its predefined modulation structure constrains the flexibility of parameter updates. These limitations motivate more expressive adaptation mechanisms that can accommodate heterogeneous control demands both across tasks and throughout individual executions.

\begin{figure*}[t]
    \centering
    \includegraphics[width=\textwidth]{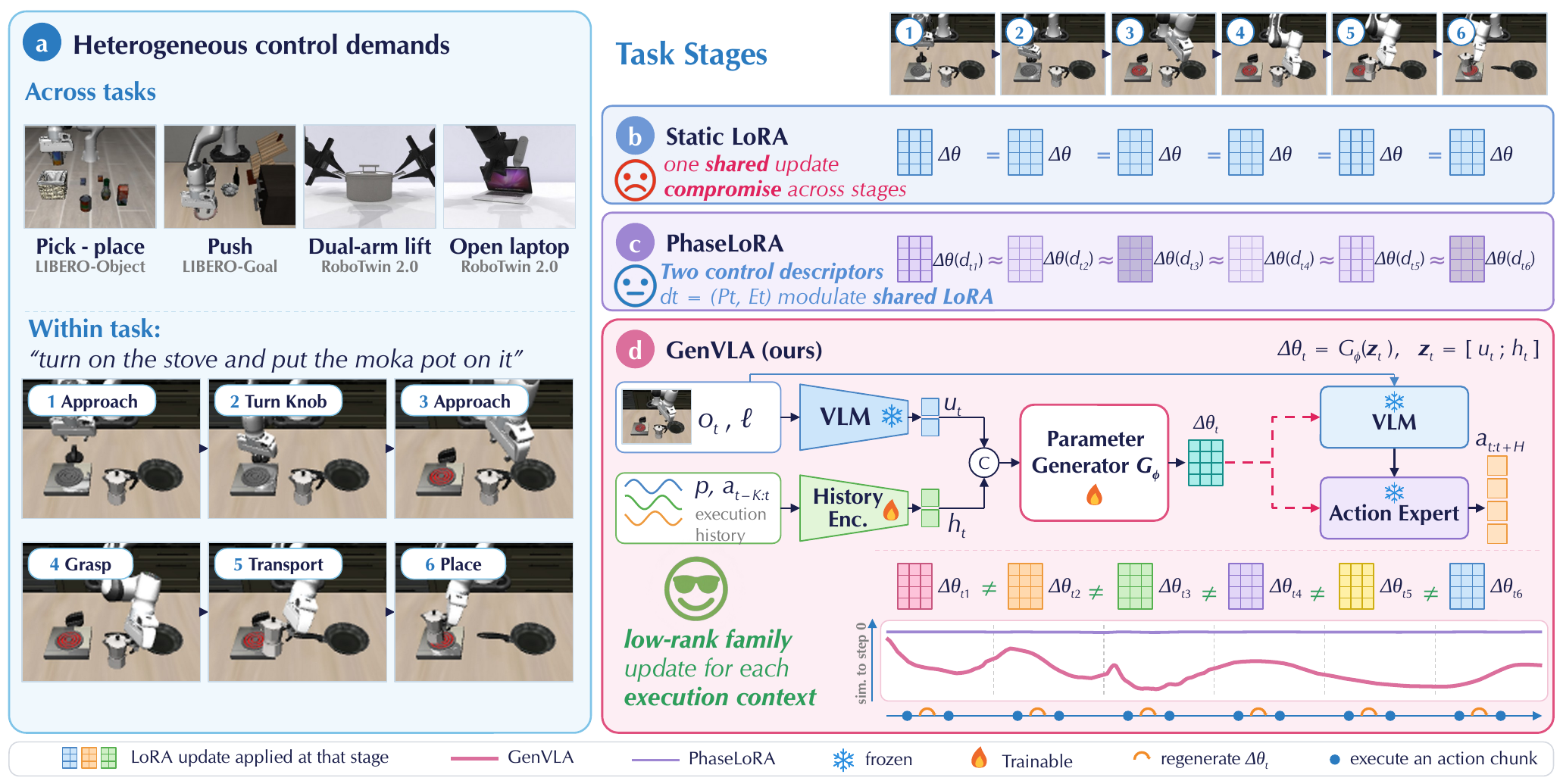}
    \caption{
    \textbf{Motivation and overview of execution-conditioned
    parameter generation.}
    (a) Heterogeneous control demands arise across
    manipulation tasks and execution stages.
    (b) Static LoRA reuses an identical parameter
    update across stages.
    (c) PhaseLoRA uses two control descriptors,
    $d_t=(P_t,E_t)$, to modulate shared LoRA
    components, yielding similar updates.
    (d) \genvla generates a diverse low-rank
    parameter family conditioned on evolving
    execution contexts.
    The lower curves show parameter-update similarity
    to the initial step in an example episode,
    highlighting greater variation in \genvla. Better viewed in color.
    }
    \label{fig:genvla_motivation}
    \vspace{-3mm}

\end{figure*}

In this work, we revisit parameter-efficient VLA adaptation from the perspective of \emph{execution-conditioned parameter generation}. Inspired by advances in hypernetworks and conditional weight adaptation~\citep{galanti2020modularity,ivison2023hint,lv2024hyperlora,ren2025hypogen,xiong2025hypervla,wu2026hy,jacob2026copra}, we propose to generate parameter updates directly from the current execution context, rather than reusing a shared update or modulating predefined components. Different contexts can thereby induce distinct low-rank updates that collectively form a diverse \emph{parameter family}. Crucially, the low-rank constraint applies to each update, not to the diversity of updates available across execution contexts. Such a parameter family can provide distinct adaptation directions for heterogeneous tasks without increasing the rank of each update. Moreover, conditioning generation on evolving states and interaction history enables the parameter family to adapt continuously to changing control demands within executions.

Motivated by this perspective, we introduce \genvla, an execution-conditioned parameter generation framework for continuous-action VLA policies. As illustrated in Fig.~\ref{fig:genvla_motivation}, a generator takes current visual-language features and short-term execution history as input and directly produces low-rank updates for a pretrained VLA's vision-language backbone (VLM) and action expert. During closed-loop execution, the generated adapter is periodically refreshed, allowing the effective policy parameters to evolve with changing control demands while all pretrained weights remain frozen. By generating distinct updates across task contexts and execution steps, \genvla constructs a \emph{dynamic low-rank parameter family} that spans diverse adaptation directions despite the extremely low rank of each update. This enables a shared policy to accommodate heterogeneous task requirements while continuously adapting to evolving control contexts. Importantly, these generated updates are not merely arbitrary temporal variations; they exhibit structured changes associated with the progression of robot manipulation. Such execution-dependent adaptation is learned in an end-to-end manner by using only the downstream action-learning objective. It requires no explicit phase annotations, predefined experts, or online gradient-based optimization.
\vspace{-1mm}

We systematically evaluate \genvla on
LIBERO~\citep{liu2023libero},
RoboTwin~2.0~\citep{chen2025robotwin},
and real-world manipulation tasks,
comparing it with static and dynamic low-rank
adaptation methods across VLA backbones.
On LIBERO, \genvla achieves \textbf{93.0\%}
average success with rank-4 generated updates,
outperforming high-rank static LoRA
(\textbf{77.1\%}) and PhaseLoRA
(\textbf{79.6\%}) under comparable
trainable-parameter budgets.
Dynamic-generation ablations demonstrate
clear benefits over fixed, global-mean,
and task-mean updates, while parameter-space
analyses reveal substantial diversity across
tasks and structured evolution throughout
individual manipulation trajectories.
Together, these findings demonstrate that
execution-conditioned parameter generation
enables expressive and adaptive low-rank
parameter families for heterogeneous robot control.

Our main contributions are summarized as follows:

\begin{itemize}
    \item We highlight across-task and within-trajectory control heterogeneity for VLA adaptation, and motivate execution-conditioned low-rank parameter families as an alternative to a single shared adaptation.
    
    \item We introduce \genvla, an execution-conditioned parameter generation framework that synthesizes diverse low-rank updates for a VLA's vision-language backbone and action expert from visual-language features and short-term execution history, enabling flexible adaptation across tasks and during closed-loop execution without phase supervision or online optimization.

    \item Extensive experiments on LIBERO,
    RoboTwin~2.0, and real-world manipulation
    tasks demonstrate the effectiveness of \genvla,
    achieving 93.0\% success on LIBERO and
    outperforming parameter-matched static LoRA
    by 15.9 points. Further analyses reveal
    diverse, execution-dependent parameter updates.
\end{itemize}

\section{Related Work}

\paragraph{Low-rank adaptation and parameter generation.}
Parameter-efficient adaptation commonly employs
LoRA~\citep{hu2021lora} and DoRA~\citep{liu2024dora},
while LoRAMoE~\citep{dou2024loramoe} and
LoRA-SP~\citep{kim2026lorasp} introduce expert
routing and adaptive capacity allocation.
These methods primarily learn shared adaptation
parameters rather than generating new updates
from the execution context.
Hypernetworks~\citep{ha2017hypernetworks}
instead predict model parameters from conditioning
signals. HINT~\citep{ivison2023hint},
HyperLoRA~\citep{lv2024hyperlora},
and Drag-and-Drop~\citep{liang2025dnd}
explore parameter generation for language models,
while LoFA~\citep{hao2026lofa},
COPRA~\citep{jacob2026copra}, and
SG-LoRA~\citep{li2026sglora}
investigate semantic- or input-conditioned
weight generation in visual and language domains.
Unlike these settings, \genvla targets closed-loop
robot control, where generated updates must respond
to evolving observations and executed actions
throughout manipulation.
\vspace{-5mm}
\paragraph{Parameter generation for robotic policies.}
Recent robotic approaches employ hypernetworks
for policy specialization.
Hyper-GoalNet~\citep{zhou2025hypergoalnet}
generates goal-conditioned policy parameters,
while DISC~\citep{ren2026disc} generates
task-specific visuomotor policies from language
instructions. HyperVLA~\citep{xiong2025hypervla}
generates compact task-specific policies
to reduce inference cost.
WIZARD~\citep{bianchi2026wizard} predicts
task-specific LoRA adapters from language
instructions and demonstration videos,
without target-task fine-tuning.
These approaches establish the effectiveness
of weight-space policy specialization,
but primarily focus on task-level adaptation
or efficient policy instantiation.
In contrast, \genvla generates low-rank
updates for a pretrained VLA from current
visual-language representations and causal
state-action history, allowing the same policy
to adapt its effective parameters repeatedly
during closed-loop execution.
\vspace{-5mm}

\paragraph{Execution-conditioned policy adaptation.}
Manipulation involves changing control demands
across approach, contact, grasping, transport,
and placement.
HyperTASR~\citep{sun2025hypertasr} uses
task and progression information to generate
adaptive scene-representation transformations,
while BehaviorVLA~\citep{hu2026behaviorvla}
learns temporally coherent behavior representations
for phase-conditioned action prediction.
PhaseLoRA~\citep{guo2026phaselora}
introduces within-trajectory adaptation by
modulating shared LoRA factors in the VLA using two control-regime descriptors.
Rather than restricting adaptation to descriptor-based modulation or intermediate representations, \genvla directly generates both low-rank factors for the VLM and action expert using visual-language features and short-term execution history. Contemporaneous work MorphAct~\citep{liu2026morphact} addresses gradient conflicts through joint VLM and action-expert adaptation, whereas \genvla focuses on evolving control demands through execution-conditioned parameter generation. This enables context-dependent updates across tasks and execution steps without explicit phase annotations or predefined experts.

\section{\genvla}
\label{sec:method}

\begin{figure*}[t]
    \centering
    \includegraphics[width=\textwidth]{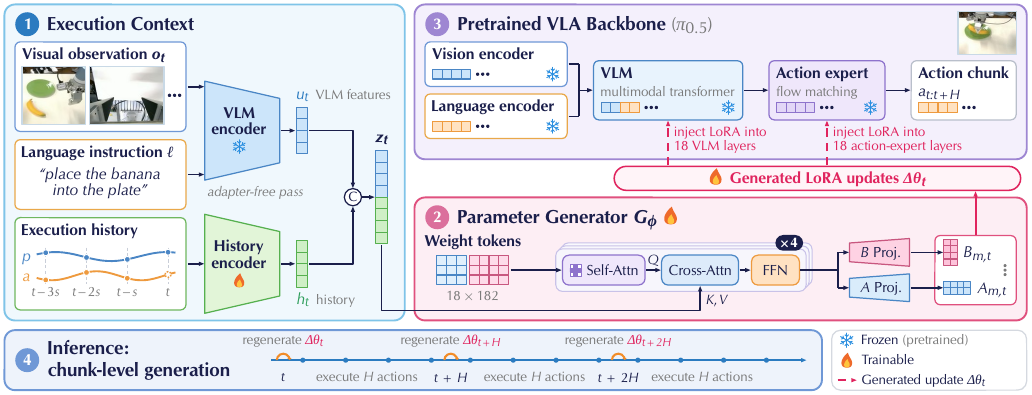}
    \caption{
    \textbf{Overview of \genvla.}
    Visual-language features and recent proprioceptive history jointly represent the execution context.
    A shared parameter generator generates context-dependent low-rank updates for the linear projections of both the VLM and the action expert.
    Across tasks and execution steps, these updates form a diverse parameter family while preserving the low rank of each update.
    The generated adapters are regenerated at every action-chunk request during closed-loop control, without explicit phase supervision or online optimization.
    }
    \label{fig:genvla_overview}
    \vspace{-5mm}
\end{figure*}

We introduce \genvla, an execution-conditioned parameter generation framework that adapts pretrained VLA policies to heterogeneous control demands across tasks and within individual executions. We instantiate \genvla on $\pi_{0.5}$~\citep{black2025pi05}, a flow-matching~\citep{lipman2023flow} VLA comprising a PaliGemma~\citep{beyer2024paligemma} vision-language backbone and an action expert. Given a visual observation $o_t$, language instruction $\ell$, and execution history $\mathcal{H}_t$, \genvla constructs an execution representation $z_t$ and generates context-dependent low-rank parameter updates:
\begin{equation}
\Delta\theta_t = G_\phi(z_t), \qquad
\pi_t = \pi_{\theta+\Delta\theta_t},
\label{eq:genvla}
\end{equation}
where $\theta$ denotes the frozen pretrained
parameters, and $G_\phi$ is a trainable
parameter generator.
Different execution contexts induce distinct
updates, enabling a shared policy to accommodate
task diversity and evolving control demands. \Cref{fig:genvla_overview} illustrates
the framework.

\subsection{Execution-Conditioned Representation}
\label{sec:context}

To capture heterogeneous control demands, \genvla constructs an execution representation from two information streams: semantic context and proprioceptive history. The former provides task and scene information, while the latter captures recent robot motion and interaction dynamics.

\stitle{Semantic Context.}
The pretrained vision-language backbone encodes the observation $o_t$ and instruction $\ell$ into hidden states $u_t\in\mathbb{R}^{L\times d}$,
where $L$ is the token length and $d$ is
the hidden dimension ($d=2048$ for the
vision-language backbone of $\pi_{0.5}$). We extract the final-layer prefix representations with a gradient-free pass of the frozen backbone in which all generated adapters are disabled, avoiding circular dependence between the generator input and its output.
After the adapters are generated, the policy re-encodes the prefix with the adapted backbone to compute the attention cache used by the action expert.
Padded language tokens are masked out of $u_t$.
These representations provide semantic and perceptual information for parameter generation.

\stitle{Execution History.}
To characterize the evolving execution state, we maintain the history of proprioceptive states and executed actions.
Given a history length $K$ and sampling stride $s$, we collect $K$ previous state-action pairs and the current state:
\vspace{-2mm}
\begin{equation}
\mathcal{H}_t =
\left[
(p_t,\mathbf{0}),
(p_{t-s},a_{t-s}^{\mathrm{exec}}),
\ldots,
(p_{t-Ks},a_{t-Ks}^{\mathrm{exec}})
\right]
\label{eq:history}
\end{equation}
where $p_t$ denotes the proprioceptive state and $a^{\mathrm{exec}}_{t-js}$ denotes the action executed from state $p_{t-js}$.
States and actions are normalized using the same
statistics as the policy inputs. For indices preceding
the episode start, proprioceptive states are clamped
to the first frame, while unavailable past actions
are zero-padded. Only previously executed actions
are included during training and closed-loop evaluation. Each history entry, formed by concatenating the state with its executed action, is embedded by a lightweight MLP (two linear layers with a GELU activation, followed by layer normalization), and a learned temporal embedding encodes its recency.
The current action is excluded, since it is the quantity the policy is about to predict; its slot in the newest entry is set to zero.
This yields $K{+}1$ history tokens $h_t$.
The VLM features are linearly projected to the generator width and concatenated with the history tokens:
\begin{equation}
z_t=[W_{\mathrm{in}}u_t;\,h_t].
\label{eq:context}
\end{equation}
The resulting representation combines task-relevant information with recent execution dynamics, allowing parameter generation to respond to both task differences and changes within a trajectory.

\subsection{Dynamic Low-Rank Parameter Generation}
\label{sec:generation}
\genvla employs a shared generator to produce
both low-rank factors conditioned on $z_t$,
enabling context-dependent parameter updates
rather than fixed adapters or modulation
of shared LoRA components.

\begin{figure}[t]
    \centering
    \includegraphics[width=0.47\textwidth]{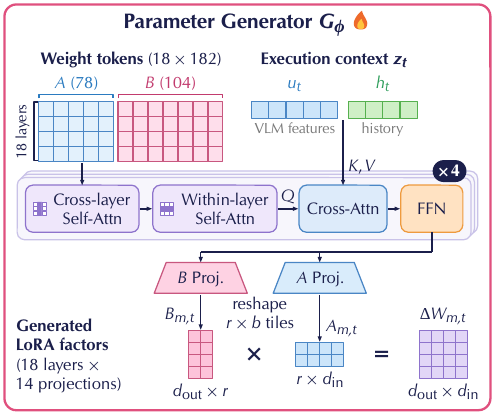}

    \caption{
        \textbf{Architecture of the \genvla parameter generator.}
        The execution context $z_t$ combines VLM features
        $u_t$ and recent execution history $h_t$.
        The $18\times182$ weight-token grid spans
        18 layer groups and 14 adapted projections,
        with 78 tokens for $A$ and 104 for $B$
        per group.
        Four attention blocks integrate weight-token
        interactions and execution context.
        Separate projections reshape the resulting
        tokens into low-rank factors $A_{m,t}$
        and $B_{m,t}$, generating context-dependent
        updates $\Delta W_{m,t}=B_{m,t}A_{m,t}$.
    }
    
    \vspace{-5mm}
    \label{fig:pg_arch}
\end{figure}

\stitle{Weight-Space Generator.}
Following the weight tokenization strategy of
\citet{wu2026hy}, we use a transformer-based
generator to produce LoRA parameters, as shown
in Fig.~\ref{fig:pg_arch}.
We partition the target LoRA factor shapes
into tiles of $r\times b$ entries, where $r$
is the LoRA rank and $b$ is the tile width
($b=512$ in our implementation).
Each tile is assigned a learnable weight token. The VLM and action expert of $\pi_{0.5}$
each contain 18 Transformer layers.
We group their weight tokens by corresponding
layer indices, using learnable embeddings
to identify the target parameter locations.
Four generator blocks process these tokens through
cross-layer and within-layer self-attention,
cross-attention to the execution context $z_t$,
and a feed-forward network.
Two separate projection heads then produce
the LoRA factors $A$ and $B$. For each adapted linear layer
$W_m\in\mathbb{R}^{d_{\mathrm{out}}\times d_{\mathrm{in}}}$,
the generator produces
$A_{m,t}\in\mathbb{R}^{r\times d_{\mathrm{in}}}$
and $B_{m,t}\in\mathbb{R}^{d_{\mathrm{out}}\times r}$:
\begin{equation}
\begin{aligned}
\{A_{m,t},B_{m,t}\}_{m\in\mathcal{S}}
    &=G_\phi(z_t),\\
\Delta W_{m,t}
    &=\frac{\alpha}{r}B_{m,t}A_{m,t},
\end{aligned}
\label{eq:generated_lora}
\end{equation}
where $\mathcal{S}$ denotes the adapted attention
and feed-forward projections of the VLM and action
expert, and $\alpha=r$
in all experiments.
The vision encoder is not adapted.

\stitle{Execution-Conditioned Parameter Family.}
A key property of \genvla is that its generated updates collectively form a context-dependent parameter family:
\begin{equation}
\mathcal{F}_\phi =
\left\{
\Delta\theta(z)\mid z\in\mathcal{Z}
\right\},
\label{eq:parameter_family}
\end{equation}
where $\mathcal{Z}$ denotes the space of execution
contexts. For each adapted layer, every generated
update satisfies
$\operatorname{rank}(\Delta W_m(z))\leq r$.
However, different contexts can induce distinct
LoRA factors and adaptation directions.
Thus, the low-rank constraint applies to each update rather than forcing all
contexts to share a fixed low-rank subspace.

This flexibility allows parameter updates to
vary across tasks and within individual
executions. Across tasks, the generator can
produce different updates for heterogeneous
manipulation demands under a shared policy.
Within an execution, the updates can change
as observations and interaction history evolve.
In both cases, each update remains low-rank,
without requiring predefined manipulation stages.

\subsection{Learning and Closed-Loop Adaptation}
\label{sec:learning}

\stitle{End-to-End Training.}
We train \genvla using the flow-matching objective
of the pretrained base VLA policy.
For each demonstration sample, the execution
representation $z_t$ is constructed from the
current observation, instruction, and history.
The generator produces $\Delta\theta_t$, and the
adapted policy predicts the velocity of the $H$-step chunk $a_{t:t+H}$.
With $\epsilon\sim\mathcal{N}(0,I)$ and
$x_\tau=\tau\epsilon+(1-\tau)a_{t:t+H}$,
the objective is
\begin{equation}
\mathcal{L}_{\mathrm{FM}}(\phi)
=
\mathbb{E}_{\tau,\epsilon}\left[
\left\|
v_{\theta+\Delta\theta_t}
(x_\tau,\tau\mid o_t,\ell)
-v_\tau^*
\right\|_2^2
\right],
\label{eq:fm_loss}
\end{equation}
where $v_\tau^*=\epsilon-a_{t:t+H}$ is the
target velocity.
The generated LoRA factors are differentiable,
allowing gradients from the action loss to
optimize the parameter generator $\phi$,
while all pretrained policy weights remain frozen.
No auxiliary phase classification, descriptor
supervision, or expert-routing losses are required.

\stitle{Closed-Loop Parameter Regeneration.}
At each replanning step, \genvla constructs
$z_t$ from the latest observation and execution
history, generates $\Delta\theta_t$, and predicts an $H$-step action chunk~\citep{zhao2023act}.
The generated adapter is reused across all
flow integration steps within the chunk.
After executing the first $n_{\mathrm{exec}}$
actions, the robot updates its history and
regenerates an adapter at the next planning step.
This allows the effective policy parameters
to change throughout execution without
online gradient optimization.
The history sampling stride $s$ determines
the temporal spacing of historical inputs,
independently of the replanning interval
$n_{\mathrm{exec}}$.

\definecolor{mypink}{RGB}{230,152,181}

\begin{table*}[t]
\centering

\caption{
\textbf{Main results on LIBERO and RoboTwin 2.0.}
Comparison with trainable-parameter-matched baselines
(approximately 8.7\%), with detailed parameter counts
provided in the supplementary material.
Success rates (\%) are averaged over three random seeds.
Avg. denotes the mean across settings, and Var.
denotes the across-seed variance in squared percentage points (pp$^2$).
In all tables, best and second-best results are
in bold and underlined, respectively.
}

\label{tab:main_results}
\vspace{-3mm}
\footnotesize
\setlength{\tabcolsep}{5pt}
\renewcommand{\arraystretch}{1.15}

\resizebox{\textwidth}{!}{%
\begin{tabular}{llccccccccccc}

\multirow{2}{*}{\textbf{\textit{Method}}}
& \multirow{2}{*}{\textbf{\textit{Source}}}
& \multirow{2}{*}{\textbf{\textit{Params. (\%)}}}
& \multicolumn{6}{c}{\textbf{LIBERO}}
& \multicolumn{4}{c}{\textbf{RoboTwin2.0}} \\

&
&
& \textbf{Spatial $\uparrow$}
& \textbf{Object $\uparrow$}
& \textbf{Goal $\uparrow$}
& \textbf{Long $\uparrow$}
& \textbf{Avg. $\uparrow$}
& \textbf{Var.}
& \textbf{C2C $\uparrow$}
& \textbf{C2R $\uparrow$}
& \textbf{Avg. $\uparrow$}
& \textbf{Var.} \\

\specialrule{0.9pt}{0pt}{2pt}

LoRA~\citep{hu2021lora}
& ICLR'22
& 8.7
& 84.1 & 91.3 & 78.2 & 54.8
& 77.1 & 2.8
& 43.2 & 25.0
& 34.1 & 3.1 \\

DoRA~\citep{liu2024dora}
& ICML'24
& 8.7
& 85.1 & 93.7 & 78.1 & 60.3
& 79.3 & 1.9
& 46.2 & 24.5
& 35.4 & 1.9 \\

LoRAMoE~\citep{dou2024loramoe}
& ACL'24
& 8.7
& 83.7 & 92.0 & 72.9 & 56.9
& 76.4 & 2.3
& 41.2 & 21.7
& 31.5 & 3.4 \\

LoFA~\citep{hao2026lofa}
& CVPR'26
& 8.7
& \underline{86.7}
& \underline{94.3}
& 79.0
& \underline{61.6}
& \underline{80.4} & 1.5
& 44.2
& \underline{27.6}
& 35.9 & 2.5 \\

PhaseLoRA~\citep{guo2026phaselora}
& NeurIPS'26
& 8.7
& 85.1 & 93.6 & \underline{81.3} & 58.2
& 79.6 & 1.9
& \underline{47.9} & 24.5
& \underline{36.2} & 2.7 \\

\specialrule{0.9pt}{3pt}{0pt}
\rowcolor{mypink!28}
\textbf{\genvla (Ours)}
& --
& 8.7
& \textbf{96.2}
& \textbf{99.0}
& \textbf{96.8}
& \textbf{80.1}
& \textbf{93.0}
& 1.2
& \textbf{59.6}
& \textbf{35.4}
& \textbf{47.5}
& 2.1 \\

\end{tabular}%
}
\vspace{-3mm}

\end{table*}
\vspace{-3mm}
\section{Experiments}
\label{sec:experiment}

We evaluate \genvla on simulated and real-world robotic
manipulation tasks to investigate three questions:
(1) whether execution-conditioned parameter generation
improves policy adaptation over existing low-rank methods;
(2) whether these improvements generalize across robotic
environments and VLA backbones; and
(3) how adaptation rank, execution history, and
dynamic parameter generation affect performance
and efficiency.

\stitle{Baselines.}
Following PhaseLoRA~\citep{guo2026phaselora}, we compare \genvla with representative
parameter-efficient adaptation methods,
including LoRA~\citep{hu2021lora},
DoRA~\citep{liu2024dora},
LoRAMoE~\citep{dou2024loramoe},
LoFA~\citep{hao2026lofa}
and PhaseLoRA~\citep{guo2026phaselora}.
All methods are jointly trained across tasks within
each benchmark, covering 40 LIBERO tasks and
50 RoboTwin 2.0 tasks, respectively.
For fair comparison, we match the trainable-parameter
budgets of all methods to approximately 8.7\%.

\stitle{Implementation Details.}
We freeze the pretrained $\pi_{0.5}$ backbone
and train only the parameter generator.
We initialize the generator to produce $B_{m,t}=0$,
ensuring zero parameter updates and preserving
the pretrained policy at initialization.
Training uses AdamW~\citep{loshchilov2019adamw} with a cosine learning-rate
schedule, linear warm-up, and gradient clipping.
We train \genvla for 50k steps on LIBERO, RoboTwin~2.0, and real-robot
data. Further implementation details are provided
in the supplementary material.

\subsection{Simulation Results}
\label{sec:sim_results}

\stitle{Evaluation Protocol.}
We evaluate all methods on 40 LIBERO tasks and
50 RoboTwin 2.0 tasks, following standard evaluation
protocols. Each LIBERO task is evaluated over
50 episodes, while each RoboTwin task is evaluated
over 100 episodes per setting.
For RoboTwin 2.0, we consider clean-to-clean (C2C)
and clean-to-randomized (C2R) evaluation, where
policies trained on clean demonstrations are tested
in clean and randomized environments.

\Cref{tab:main_results} shows the
results on LIBERO and RoboTwin 2.0.
\ding{182} On LIBERO, \genvla achieves an average success
rate of 93.0\%, outperforming the evaluated
static and dynamically adapted LoRA baselines.
Improvements are observed across all 4 suites,
including the challenging long-horizon tasks,
where \genvla achieves 80.1\% success rate.

\noindent \ding{183} On RoboTwin 2.0, \genvla achieves 59.6\% and
35.4\% success rate under C2C and C2R settings,
respectively.
These results demonstrate that execution-conditioned
parameter generation is effective across different benchmarks and remains beneficial
under visual distribution shifts.

\begin{figure}[h]
  \centering
  \includegraphics[width=\columnwidth]{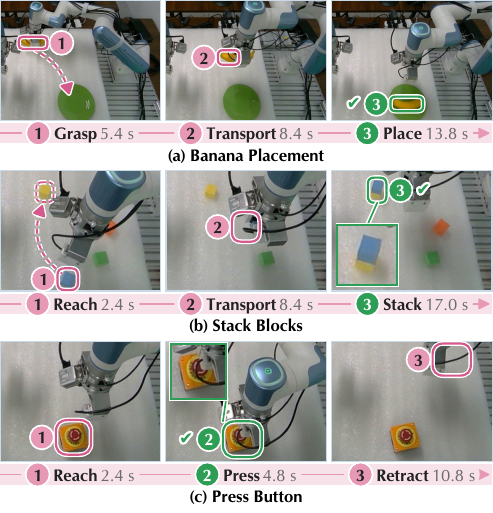}
  \caption{\textbf{Real-robot tasks.} Each panel shows three frames of one demonstration
  in temporal order (left to right): (a) grasp, carry, and place the banana;
  (b) grasp the blue block and stack it on the yellow one; (c) approach and press the button.}
  \label{fig:real_world}
\end{figure}

\subsection{Real-World Evaluation}
\label{sec:real_robot}

We further evaluate \genvla on three real-world
manipulation tasks (\cref{fig:real_world}):
banana pick-and-place, block stacking, and
button pressing, with 20 trials per task.
As shown in Table~\ref{tab:real_robot},
\genvla achieves success rates of 100\%,
80\%, and 70\%, respectively.
These results demonstrate the feasibility
of execution-conditioned parameter generation
in physical manipulation.
Additional experimental details and analysis
are provided in the supplementary material.

\begin{table}[t]
\centering
\caption{
\textbf{Real-world manipulation success rates.}
Each method is evaluated over 20 trials per task.}
\label{tab:real_robot}

\footnotesize
\setlength{\tabcolsep}{5pt}
\renewcommand{\arraystretch}{1.15}

\begin{tabularx}{\columnwidth}{
l
*{3}{>{\centering\arraybackslash}X}
}

\textbf{\textit{Method}}
& \textbf{Banana $\uparrow$}
& \textbf{Blocks $\uparrow$}
& \textbf{Button $\uparrow$} \\

\specialrule{0.9pt}{0pt}{2pt}

Static LoRA
& 60\%
& 5\%
& 0\% \\

Full Fine-tuning
& 75\%
& 0\%
& 0\% \\

\specialrule{0.9pt}{3pt}{0pt}

\rowcolor{mypink!28}
\textbf{\genvla (Ours)}
& \textbf{100\%}
& \textbf{80\%}
& \textbf{70\%} \\

\end{tabularx}

\end{table}

\begin{table}[t]
\centering
\caption{
\textbf{Cross-backbone generalization on LIBERO using X-VLA.}
Success rates (\%) are reported across 4 task suites.
}
\label{tab:xvla_generalization}

\footnotesize
\setlength{\tabcolsep}{5pt}
\renewcommand{\arraystretch}{1.15}

\resizebox{\columnwidth}{!}{%
\begin{tabular}{lccccc}

\textbf{\textit{Method}}
& \textbf{Spatial $\uparrow$}
& \textbf{Object $\uparrow$}
& \textbf{Goal $\uparrow$}
& \textbf{Long $\uparrow$}
& \textbf{Avg. $\uparrow$} \\

\specialrule{0.9pt}{0pt}{2pt}

Static LoRA
& 83 & 86 & 82 & 56 & 76.8 \\

\specialrule{0.9pt}{3pt}{0pt}

\rowcolor{mypink!28}
\textbf{\genvla (X-VLA)}
& \textbf{93}
& \textbf{99}
& \textbf{83}
& \textbf{79}
& \textbf{88.5} \\

\end{tabular}%
}

\end{table}

\begin{table}[t]
\centering
\caption{
\textbf{Architecture and adaptation ablations on LIBERO.}
Success rates (\%) are reported across
four task suites. All variants use rank-$4$.
Params. denotes the percentage of trainable parameters.
}
\label{tab:pg_arch}

\footnotesize
\setlength{\tabcolsep}{4pt}
\renewcommand{\arraystretch}{1.15}

\begin{tabularx}{\columnwidth}{
l
>{\raggedright\arraybackslash}X
>{\centering\arraybackslash}p{0.20\columnwidth}
>{\centering\arraybackslash}p{0.11\columnwidth}
}

\textbf{\textit{Factor}}
& \textbf{Setting}
& \textbf{Params. (\%)}
& \textbf{Avg. $\uparrow$} \\

\specialrule{0.9pt}{0pt}{2pt}

\multirow{3}{*}{Generator}
& 2L, d2048
& 4.67 & 89.0 \\

& \cellcolor{mypink!28}4L, d2048 (default)
& \cellcolor{mypink!28}8.71
& \cellcolor{mypink!28}\textbf{93.8} \\

& 4L, d4096
& 27.48 & 90.0 \\

\midrule

\multirow{2}{*}{Compact Gen.}
& Small MLP
& 0.35 & 79.5 \\
& Small Transformer
& 0.35 & \textbf{85.8} \\

\midrule

\multirow{2}{*}{Factor Gen.}
& B-only
& 8.77 & 82.8 \\

& \cellcolor{mypink!28}A+B (default)
& \cellcolor{mypink!28}8.71
& \cellcolor{mypink!28}\textbf{93.8} \\

\specialrule{\lightrulewidth}{0pt}{\belowrulesep}
\multirow{3}{*}{Placement}
& Action only
& 8.70 & 54.5 \\

& VLM only
& 8.71 & 85.8 \\

& \cellcolor{mypink!28}VLM + Action (default)
& \cellcolor{mypink!28}8.71
& \cellcolor{mypink!28}\textbf{93.8} \\

\end{tabularx}

\end{table}

\subsection{Cross-Backbone Evaluation}
\label{sec:cross_backbone}

To examine whether \genvla generalizes beyond
a specific VLA architecture, we integrate the
parameter generator into X-VLA~\citep{zheng2025xvla},
adapting the parameter-matched updates to its
architecture and training for 50k steps.
As shown in Table~\ref{tab:xvla_generalization},
it improves four suites over static LoRA
(88.5\% on average).

\subsection{Extended Adaptation}
\label{sec:extended_adaptation}

While our main experiments focus on frozen-backbone
adaptation, \genvla is compatible with broader fine-tuning
strategies.
We further investigate different adaptation scopes on LIBERO,
including a comparison with MorphAct~\citep{liu2026morphact},
a concurrent parameter generation method.
As shown in \Cref{tab:adaptation_scope}, when training only
the parameter generators, \genvla outperforms MorphAct
by 15.3 percentage points.
Additionally optimizing the vision tower and action/time
interface further improves \genvla from 93.8\% to 99.5\%,
compared with 98.5\% for MorphAct, demonstrating its
effectiveness across different adaptation scopes.

\begin{table}[t]
\centering
\caption{
\textbf{Extended adaptation on LIBERO.}
Success rates (\%) are evaluated over 400 episodes
with a fixed seed.
Trainable parameters are reported in millions.
$\dagger$ denotes our reproduction using the official codebase.
}
\label{tab:adaptation_scope}

\footnotesize
\setlength{\tabcolsep}{5pt}
\renewcommand{\arraystretch}{1.15}

\resizebox{\columnwidth}{!}{%
\begin{tabular}{lccccc}

\multirow{2}{*}{\textbf{\textit{Method}}}
& \multicolumn{4}{c}{\textbf{Trainable Params. (M)}}
& \multirow{2}{*}{\textbf{Avg. $\uparrow$}} \\

\cmidrule(lr){2-5}

& \textbf{Gen.}
& \textbf{Vision}
& \textbf{Interface}
& \textbf{Total}
& \\

\specialrule{0.9pt}{0pt}{2pt}

MorphAct (Frozen)$^\dagger$
& 300.09 & 0 & 0 & 300.09 & 78.5 \\

\rowcolor{mypink!28}
\genvla (Frozen)
& 345.07 & 0 & 0 & 345.07 & 93.8 \\

\specialrule{0.9pt}{0pt}{2pt}

MorphAct (Extended)$^\dagger$
& 300.09 & 414.80 & 2.17 & 717.06
& \underline{98.5} \\

\rowcolor{mypink!28}
\textbf{\genvla (Extended)}
& 345.07 & 414.80 & 2.17 & 762.04
& \textbf{99.5} \\

\end{tabular}%
}

\end{table}

\subsection{Adaptation Efficiency}
\label{sec:adaptation efficiency}

\begin{figure}[h]
    \centering
    \includegraphics[width=1\columnwidth]
    {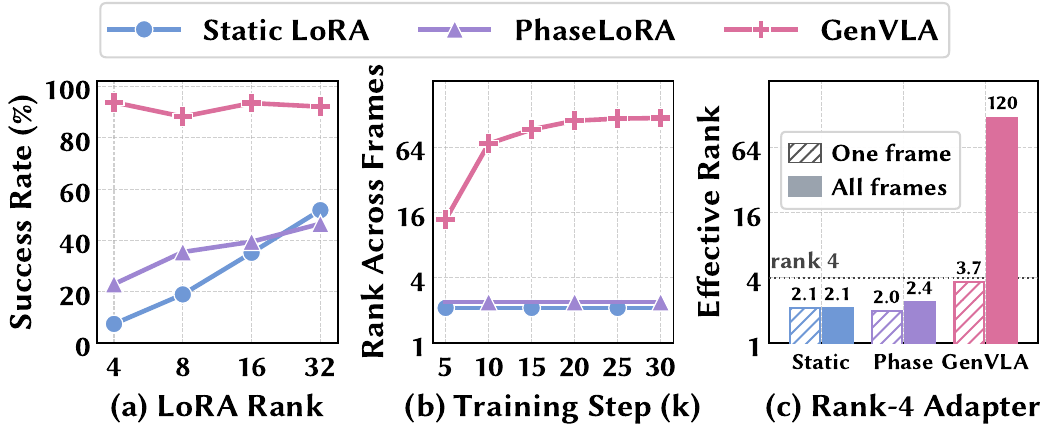}
    \caption{
    \textbf{Effect of LoRA rank on LIBERO.}
    (a) Success rate versus rank over 400 evaluation
    episodes per configuration.
    (b) Effective rank of rank-4 updates across
    frames during training.
    (c) Effective rank of one frame versus all frames.
    }
    \label{fig:rank_sweep}
    
\end{figure}

\stitle{Inference Cost.}
\genvla generates low-rank updates once per
replanning context and reuses them throughout
action-chunk prediction.
Under identical inference settings on an NVIDIA
A100 GPU and matched rank ($r=4$),
\genvla requires 527.0 ms per action chunk,
compared with 460.7 ms for static LoRA
and 767.4 ms for PhaseLoRA.
This corresponds to a 14.4\% latency overhead
over static LoRA, while reducing latency
by 31.3\% relative to PhaseLoRA.

\subsection{Ablation Studies and Analysis}
\label{sec:ablation}
We use a fixed seed for the following ablations
(93.8\% vs. 93.0\% averaged over three seeds
in the main table), examining adapter
expressivity, conditioning and adaptation
mechanisms, and temporal dynamics and
functional alignment during robot execution.

\stitle{Low-Rank Adaptation Analysis.}
As shown in Fig.~\ref{fig:rank_sweep}(a),
\genvla outperforms static LoRA and PhaseLoRA
across ranks, with the largest gap at $r=4$.
Its performance remains stable from $r=4$ to $r=32$, achieving 93.8\% at $r=4$ for a fixed seed.
All three methods constrain each per-frame
update to rank at most $r$.
Across execution frames, however, static LoRA
retains fixed updates, PhaseLoRA dynamically
modulates shared low-rank factors, while
\genvla generates execution-conditioned factors,
forming a family of adapters spanning
diverse parameter directions.
For dynamic adaptation methods, combining
updates across frames can yield a matrix
with a much higher rank than individual updates.
We quantify this cross-frame diversity using
effective rank, which measures the number of
meaningfully contributing directions based on
the singular values of concatenated updates
(see the supplementary material).
As shown in Fig.~\ref{fig:rank_sweep}(b,c),
the effective rank increases during training,
reaching approximately $120$ at rank $4$,
compared with $2.1$ for static LoRA
and $2.4$ for PhaseLoRA.
These results highlight the greater cross-frame
diversity of \genvla's low-rank adapter family. Given its stable performance across ranks,
we adopt $r=4$ as the default configuration
to reduce the size of generated updates
without sacrificing accuracy.

\stitle{Architecture and Adaptation.}
As shown in Table~\ref{tab:pg_arch},
the default generator achieves the best performance
among architecture variants, while increasing
its width or reducing its depth degrades accuracy.
Notably, a compact Transformer generator with only 0.35\%
trainable parameters achieves 85.8\% success,
outperforming a parameter-matched MLP (79.5\%),
suggesting effective parameter generation
even with limited capacity.
Generating both LoRA factors and adapting
both the VLM and action expert improve performance.

\begin{table}[t]
\centering
\caption{
\textbf{Conditioning ablation on LIBERO.}
Success rates (\%) of trained
variants with selected generator inputs removed.
}
\label{tab:conditioning_ablation}

\footnotesize
\setlength{\tabcolsep}{5pt}
\renewcommand{\arraystretch}{1.15}

\resizebox{\columnwidth}{!}{%
\begin{tabular}{lccccc}

\textbf{\textit{Condition}}
& \textbf{Spatial $\uparrow$}
& \textbf{Object $\uparrow$}
& \textbf{Goal $\uparrow$}
& \textbf{Long $\uparrow$}
& \textbf{Avg. $\uparrow$} \\

\specialrule{0.9pt}{0pt}{2pt}

w/o Visual Token
& 91 & 96 & \underline{93} & 75 & \underline{88.8} \\

w/o Language Token
& 91 & \underline{98} & 86 & 73 & 87.0 \\

w/o Action History
& 87 & 93 & 92
& \underline{76} & 87.0 \\

w/o Proprio. History
& \underline{94} & 94 & 88 & 59 & 83.8 \\

w/o Both History
& 89 & 91 & 88 & 63 & 82.8 \\

\specialrule{0.9pt}{3pt}{0pt}

\rowcolor{mypink!28}
\textbf{Full \genvla}
& \textbf{97}
& \textbf{99}
& \textbf{99}
& \textbf{80}
& \textbf{93.8} \\

\end{tabular}%
}

\end{table}

\begin{table}[t]
\centering
\caption{
\textbf{History-conditioned adaptation on LIBERO.}
Success rates (\%) with identical conditioning
inputs and comparable trainable parameter budgets.
}
\label{tab:history_adaptation}

\footnotesize
\setlength{\tabcolsep}{5pt}
\renewcommand{\arraystretch}{1.15}

\begin{tabularx}{\columnwidth}{
l
*{2}{>{\centering\arraybackslash}X}
}

\textbf{\textit{Method}}
& \textbf{Params. (\%) }& \textbf{Avg. (\%) $\uparrow$} \\

\specialrule{0.9pt}{0pt}{2pt}

History-RGate
& 8.84 & 74.8 \\

History-FiLM
& 8.85 & \underline{85.5} \\

\specialrule{0.9pt}{3pt}{0pt}

\rowcolor{mypink!28}
\textbf{\genvla (Ours)}
& 8.71
& \textbf{93.8} \\

\end{tabularx}
\end{table}

\begin{table}[t]
\centering
\caption{
\textbf{Effect of temporal parameter regeneration
on LIBERO.}
Success rates (\%) with different adapter update
schedules. PG denotes parameter generation.
}
\label{tab:temporal_regeneration}

\footnotesize
\setlength{\tabcolsep}{5pt}
\renewcommand{\arraystretch}{1.15}

\resizebox{\columnwidth}{!}{%
\begin{tabular}{lccccc}

\textbf{\textit{Adapter}}
& \textbf{Spatial $\uparrow$}
& \textbf{Object $\uparrow$}
& \textbf{Goal $\uparrow$}
& \textbf{Long $\uparrow$}
& \textbf{Avg. $\uparrow$} \\

\specialrule{0.9pt}{0pt}{2pt}

Episode-Level PG
& \underline{86}
& 92
& 83
& 56
& 79.3 \\

Phase-Based PG
& 81
& \underline{97}
& \underline{84}
& \underline{69}
& \underline{82.8} \\

\specialrule{0.9pt}{3pt}{0pt}

\rowcolor{mypink!28}
\textbf{\genvla (Per-Replan)}
& \textbf{97}
& \textbf{99}
& \textbf{99}
& \textbf{80}
& \textbf{93.8} \\

\end{tabular}%
}

\end{table}

\begin{table}[t]
\centering
\caption{
\textbf{Inference-time adapter interventions on LIBERO.}
Success rates (\%) over 400 episodes.
Fixed and mismatched adapters replace
the generated updates without retraining.
}
\label{tab:dynamic_ablation}

\footnotesize
\setlength{\tabcolsep}{5pt}
\renewcommand{\arraystretch}{1.15}

\resizebox{\columnwidth}{!}{%
\begin{tabular}{lccccc}

\textbf{\textit{Adapter}}
& \textbf{Spatial $\uparrow$}
& \textbf{Object $\uparrow$}
& \textbf{Goal $\uparrow$}
& \textbf{Long $\uparrow$}
& \textbf{Avg. $\uparrow$} \\

\specialrule{0.9pt}{0pt}{2pt}

Global Mean
& \underline{45} & \underline{51} & 47 & 5 & 37.0 \\

Task Mean
& \underline{45} & 48 & \underline{61}
& \underline{12} & \underline{41.5} \\

Frozen
& 10 & 0 & 29 & 0 & 9.8 \\

Shuffled
& 26 & 1 & 27 & 1 & 13.8 \\

\specialrule{0.9pt}{3pt}{0pt}

\rowcolor{mypink!28}
\textbf{\genvla (Dynamic)}
& \textbf{97}
& \textbf{99}
& \textbf{99}
& \textbf{80}
& \textbf{93.8} \\

\end{tabular}%
}

\end{table}

\stitle{Conditioning Signals.}
\genvla conditions parameter generation on current
visual-language features and recent proprioceptive
and action histories.
To assess their contributions, we train
ablated variants with selected conditioning inputs
removed, while preserving the policy inputs.
As shown in Table~\ref{tab:conditioning_ablation},
removing visual features reduces success from
93.8\% to 88.8\%, while removing language features
or action history yields 87.0\%.
Excluding proprioceptive history causes a drop to 83.8\%, with removing both histories
yielding 82.8\%.
The impact is particularly pronounced on
LIBERO-Long, where success drops from 80\%
to 59\% without proprioceptive history.
These results highlight the importance of
execution history, particularly proprioceptive
information, for adaptive parameter generation.

\begin{figure}[!t]
    \centering
    \includegraphics[width=\columnwidth]
{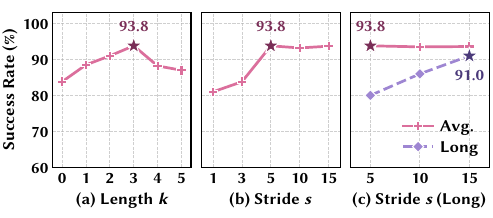}
    \caption{
    \textbf{Effect of execution history on LIBERO.}
    We vary the history length $K$ and sampling
    stride $s$, adopting $K=3$ and $s=5$
    for the final configuration.
    }
    \label{fig:history}

\end{figure}

\stitle{Temporal Sensitivity.}
We investigate the effects of history
length $K$ and sampling stride $s$.
As shown in Fig.~\ref{fig:history}(a--b),
\genvla achieves 93.8\% average success rate
at $K=3$ and $s=5$, while longer
histories reduce performance.
Increasing the stride from $5$ to $15$
improves LIBERO-Long success from 80\%
to 91\%, while maintaining comparable overall
performance (Fig.~\ref{fig:history}(c)).
This suggests that broader temporal coverage
can benefit long-horizon manipulation.

\stitle{History-Conditioned Adaptation.}
To test whether execution history alone
explains the gains, we compare \genvla with
history-conditioned FiLM~\citep{perez2018film}
and normalized rank-wise gating
under matched inputs and comparable parameter
budgets. FiLM modulates intermediate features,
while gating reweights shared low-rank factors.
Both underperform \genvla (85.5\% and 74.8\%
vs. 93.8\%), supporting the benefits of
context-specific weight generation
(Tab.~\ref{tab:history_adaptation}).

\stitle{Temporal Regeneration.}
We compare generators trained and evaluated
under different update schedules.
Episode-Level PG generates adapters once per
episode, while Phase-Based PG regenerates them
at gripper-command transitions (open/close).
Both reuse cached adapters between updates.
Compared with these variants, per-replan
\genvla achieves higher success
(Tab.~\ref{tab:temporal_regeneration}),
supporting continuous execution-conditioned
parameter adaptation.

\stitle{Dynamic Generation.}
To examine whether \genvla
collapses to static adaptation, we perform
inference-time interventions.
Global Mean and Task Mean replace dynamic updates
with fixed averages, Frozen reuses the initial
adapter, and Shuffled uses mismatched adapters.
All degrade performance
(37.0\%, 41.5\%, 9.8\%, and 13.8\%,
respectively) relative to \genvla (93.8\%)
(Tab.~\ref{tab:dynamic_ablation}).
These results indicate reliance on
context-aligned parameter updates.

\begin{figure}[!t]
    \centering
    \includegraphics[width=1\columnwidth]
    {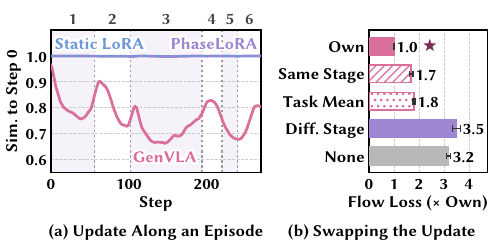}

    \caption{
    \textbf{Temporal dynamics and functional relevance
    of policy adapters.}
    (a) Cosine similarity of parameter
    updates to the initial update along a
    LIBERO-Long demonstration.
    Numbers indicate six manipulation stages:
    approach, knob turning, object approach,
    grasp, transport, and placement.
    (b) Relative flow-matching loss under
    adapter replacement on held-out LIBERO
    demonstrations.
    Same- and different-stage replacements
    are sampled across demonstrations
    without matching temporal progress.
    Lower is better.
    }

    \label{fig:execution_alignment}
\end{figure}

\begin{table}[t]
\centering
\caption{
\textbf{Temporal dynamics and functional
sensitivity on LIBERO.}
Temporal statistics use 400 closed-loop episodes
per method; adapter swapping uses 480 held-out
targets per method.
Temporal metrics are averaged across four suites.
Flow loss is normalized to each model's own adapter.
}
\label{tab:execution_alignment}

\footnotesize
\setlength{\tabcolsep}{4pt}
\renewcommand{\arraystretch}{1.12}

\begin{tabularx}{\columnwidth}{@{}Xcc@{}}
\textbf{\textit{Metric}}
& \textbf{\genvla}
& PhaseLoRA \\

\specialrule{0.9pt}{0pt}{2pt}
{
\textit{Temporal Variation}
} \\

Direction change ($1-\cos$)
& 0.024 & 0.00019 \\

Magnitude change
& 0.017 & 0.041 \\

Episode-end drift ($1-\cos$)
& 0.21 & 0.0011 \\

\midrule
\multicolumn{3}{l}{
\textit{Event Alignment}
} \\

Gripper-event enrichment
& 3.01 & 3.75 \\

\midrule
\multicolumn{3}{l}{
\textit{Adapter Swapping (Relative Flow Loss)}
} \\

Nearby ($\pm4$ steps)
& 1.15 & 1.10 \\

Same stage, 8 steps
& 1.52 & 1.23 \\

Same stage, 16 steps
& 2.61 & 1.50 \\

Same stage, 32 steps
& 3.55 & 1.74 \\

\bottomrule
\end{tabularx}

\end{table}

\stitle{Execution Alignment.}
We analyze the temporal dynamics and functional
relevance of generated adapters.
\Cref{fig:execution_alignment}(a) illustrates substantial directional changes
in \genvla, whereas PhaseLoRA largely preserves
its update direction.
Across 400 closed-loop LIBERO episodes per method,
Table~\ref{tab:execution_alignment} confirms
that \genvla exhibits greater directional variation,
while PhaseLoRA primarily modulates update magnitudes.
Both methods show changes aligned with gripper events. To assess functional relevance, we replace
the current adapter with updates from other
execution contexts.
As illustrated in Fig.~\ref{fig:execution_alignment}(b),
mismatched adapters increase
the flow-matching loss.
Controlled comparisons further show that
even within the same manipulation stage,
adapters from 16--32 steps away increase
the loss to 2.61--3.55$\times$.
These results support temporally evolving,
execution-context-sensitive adaptation
beyond coarse stage condition.
\vspace{-6mm}
\section{Conclusion}

We presented \genvla, an execution-conditioned
parameter generation framework that addresses
heterogeneous control demands both across tasks
and within individual executions.
By generating context-specific low-rank updates,
\genvla constructs a diverse parameter family
while preserving the low rank of each update.
Experiments on LIBERO, RoboTwin~2.0, and real
robots demonstrate its effectiveness, achieving
93.0\% success on LIBERO with rank-$4$ updates.
Ablations reveal the importance of dynamic
generation and execution history, while analyses
of parameter diversity and execution alignment
support context-specific adaptation across tasks
and within individual executions.

{
    \small
    \bibliographystyle{ieeenat_fullname}
    \bibliography{main}
}


\appendix

\section{Baseline Implementation Details}
\label{app:baseline_details}

\subsection{Baseline Configurations and Fairness.}

Following the controlled comparison protocol of
PhaseLoRA~\citep{guo2026phaselora},
we evaluate representative PEFT methods,
including LoRA~\citep{hu2021lora},
DoRA~\citep{liu2024dora},
LoRAMoE~\citep{dou2024loramoe},
LoFA~\citep{hao2026lofa},
and PhaseLoRA~\citep{guo2026phaselora}.
As LoFA was originally designed for visual
generative models, we adapt its Stage-II
LoRA predictor to the VLA backbone and train
it with the action flow-matching objective.
To isolate the contribution of execution-conditioned
parameter generation, we use consistent backbones,
training data, and optimization settings while
matching trainable-parameter budgets to approximately
8.7\%, including method-specific components,
as detailed in Table~\ref{tab:baseline_configs}.
All baselines are trained and evaluated
under our unified protocol, with detailed
configurations provided in
\Cref{tab:baseline_params}.

\subsection{Training and Evaluation Protocol.}
\label{tab:baseline_params}
All methods are trained using AdamW with a batch
size of 32 for 50,000 optimization steps.
The learning rate is linearly warmed up for
1,000 steps to $10^{-5}$, followed by cosine
decay to $10^{-6}$.
We jointly train across 40 LIBERO tasks and
50 RoboTwin 2.0 tasks, respectively.
The action-chunk horizons are set to 10 for
LIBERO and 50 for RoboTwin 2.0.
For the main results, each LIBERO task is evaluated
over 50 episodes, while each RoboTwin 2.0 task
is evaluated over 100 episodes under both
clean-to-clean (C2C) and clean-to-randomized
(C2R) settings.
All main results are averaged over three random seeds.
For computational efficiency, ablation studies and
supplementary experiments use a single fixed seed,
with 10 episodes per LIBERO task following
PhaseLoRA~\citep{guo2026phaselora},
unless otherwise specified.

\begin{table*}[t]
\centering
\caption{
\textbf{Parameter-matched baseline configurations.}
Adaptation ranks, LoRA scaling factors ($\alpha$),
and exact trainable-parameter counts on LIBERO and
RoboTwin 2.0. Params. denotes the percentage of
trainable parameters. For \genvla{}, the 8.71\% reported in Table~1 of the main paper
is the fraction of trainable parameters on LIBERO ($r=4$).
On RoboTwin~2.0, the higher adaptation rank ($r=16$)
increases this fraction slightly to 8.74\%.
}
\label{tab:baseline_configs}

\footnotesize
\setlength{\tabcolsep}{5pt}
\renewcommand{\arraystretch}{1.15}

\resizebox{\textwidth}{!}{%
\begin{tabular}{lcccccc}
\toprule

\multirow{2}{*}{\textbf{Method}}
& \multicolumn{3}{c}{\textbf{LIBERO}}
& \multicolumn{3}{c}{\textbf{RoboTwin 2.0}} \\

\cmidrule(lr){2-4}
\cmidrule(lr){5-7}

& \textbf{Rank / $\alpha$}
& \textbf{Trainable Params.}
& \textbf{Params. (\%)}
& \textbf{Rank / $\alpha$}
& \textbf{Trainable Params.}
& \textbf{Params. (\%)} \\

\midrule

LoRA~\citep{hu2021lora}
& 208 / 208
& 345,047,040
& 8.71
& 208 / 208
& 345,047,040
& 8.71 \\

DoRA~\citep{liu2024dora}
& 207 / 207
& 344,328,192
& 8.69
& 207 / 207
& 344,328,192
& 8.69 \\

LoRA-MoE~\citep{dou2024loramoe}
& $4\times52$ / 52
& 346,115,076
& 8.73
& $4\times52$ / 52
& 346,115,076
& 8.73 \\

LoFA~\citep{hao2026lofa}
& 4 / 4
& 344,353,429
& 8.69
& 4 / 4
& 344,353,429
& 8.69 \\

PhaseLoRA~\citep{guo2026phaselora}
& 147 / 147
& 345,833,602
& 8.73
& 147 / 147
& 345,833,602
& 8.73 \\

\midrule

\textbf{GenVLA (Ours)}
& 4 / 4
& 345,067,520
& 8.71
& 16 / 16
& 346,212,352
& 8.74 \\

\bottomrule
\end{tabular}%
}
\end{table*}

\subsection{Adaptation Rank Selection and Comparison.}
We adopt benchmark-specific adaptation ranks
to balance effectiveness and efficiency.    
On LIBERO, \genvla achieves the close average
success rate of 93.8\% and 93.5\%  at ranks 4 and 16, respectively. We select $r=4$ to reduce the size of
generated updates.
On RoboTwin 2.0, increasing the rank from
4 to 16 improves C2C success from 50.0\% (tested by three seeds)
to 59.6\%, motivating the use of $r=16$.
We use rank 4 by default for LIBERO ablations unless rank is varied explicitly. To complement the parameter-matched comparisons,
we further evaluate all methods at fixed ranks
of $r=4$ and $r=16$ on LIBERO.
As identical ranks do not imply equal
trainable-parameter budgets across methods,
we report both trainable-parameter percentages and success
rates in \Cref{tab:rank_controlled}.

\begin{table}[t]
\centering
\caption{
\textbf{Rank-controlled comparison on LIBERO.}
Average success rates (\%) are evaluated over
400 episodes using a fixed seed.
Params. denotes the percentage of trainable parameters. LoRAMoE uses four experts, each with the
specified rank.
}
\label{tab:rank_controlled}

\footnotesize
\setlength{\tabcolsep}{3.5pt}
\renewcommand{\arraystretch}{1.15}

\begin{tabular}{lcccc}
\toprule

\multirow{2}{*}{\textbf{Method}}
& \multicolumn{2}{c}{$r=4$}
& \multicolumn{2}{c}{$r=16$} \\

\cmidrule(lr){2-3}
\cmidrule(lr){4-5}

& Params. (\%)
& Avg. $\uparrow$
& Params. (\%)
& Avg. $\uparrow$ \\

\midrule

LoRA~\citep{hu2021lora}
& 0.18 & 7.5
& 0.73 & 35.2 \\

DoRA~\citep{liu2024dora}
& 0.21 & 21.5
& 0.75 & 40.5 \\

LoRAMoE~\citep{dou2024loramoe}
& 0.76 & 18.0
& 2.88 & 33.8 \\

LoFA~\citep{hao2026lofa}
& 8.69 & 80.8
& 25.68 & 86.0 \\

PhaseLoRA~\citep{guo2026phaselora}
& 0.27 & 23.0
& 1.04 & 39.5 \\

\midrule

\textbf{GenVLA (Ours)}
& 8.71 & \textbf{93.8}
& 8.74 & \textbf{93.5} \\
\bottomrule
\end{tabular}
\end{table}

\section{Comparison with Concurrent Parameter Generation Methods}
\label{app:concurrent}

\subsection{Methodological Differences.}
MorphAct~\citep{liu2026morphact} employs two
coupled generators, where the action generator
is conditioned on features from the adapted VLM.
In contrast, \genvla uses a shared generator
conditioned on visual-language features and
recent state-action history to jointly generate
VLM and action-expert updates.
While both methods produce context-dependent
low-rank adaptations, MorphAct emphasizes
representation coupling, whereas \genvla
explicitly models evolving execution history. 

\subsection{Comparison Across Adaptation Settings.}
We compare both methods using the same
$\pi_{0.5}$ backbone and LIBERO evaluation
protocol, with results summarized in
\Cref{tab:concurrent_pg}.
MorphAct is implemented using its official code.
We first consider a frozen-backbone setting,
where only the parameter generators are trained.
Under this setting, \genvla achieves 93.8\%
success, outperforming MorphAct (78.5\%)
by 15.3 percentage points.
We further adopt the released MorphAct
fine-tuning scope, additionally training
the vision tower and action/time interface.
Under this setting, \genvla achieves 99.5\%
success, compared with 98.5\% for MorphAct.
These results support the effectiveness of
execution-conditioned parameter generation
under both adaptation settings.

\begin{table}[t]
\centering
\caption{
\textbf{Comparison with MorphAct on LIBERO.}
Average success rates (\%) are evaluated over
400 episodes with a fixed seed.
Trainable parameters are reported in millions.
Gen. denotes generator parameters, while Vision
and Interface denote directly optimized
vision-tower and action/time interface parameters.
The frozen variants train only their generators.
$\dagger$ denotes our reimplementation
using the official MorphAct codebase.
}
\label{tab:concurrent_pg}

\footnotesize
\setlength{\tabcolsep}{3.5pt}
\renewcommand{\arraystretch}{1.15}

\resizebox{\columnwidth}{!}{%
\begin{tabular}{lccccc}
\toprule

\multirow{2}{*}{\textbf{Method}}
& \multicolumn{4}{c}{\textbf{Trainable Params. (M)}}
& \multirow{2}{*}{\textbf{Avg. $\uparrow$}} \\

\cmidrule(lr){2-5}

& \textbf{Gen.}
& \textbf{Vision}
& \textbf{Interface}
& \textbf{Total}
& \\

\midrule

MorphAct (Frozen)$^\dagger$
& 300.09 & 0 & 0 & 300.09 & 78.5 \\

\textbf{GenVLA (Frozen)}
& 345.07 & 0 & 0 & 345.07
& \textbf{93.8} \\

\midrule

MorphAct (Released)$^\dagger$
& 300.09 & 414.80 & 2.17
& 717.06 & 98.5 \\

\textbf{GenVLA (Released-style)}
& 345.07 & 414.80 & 2.17
& 762.04 & \textbf{99.5} \\

\bottomrule
\end{tabular}%
}
\end{table}

\section{Inference Latency Analysis}
\label{app:inference}

\noindent\textbf{Inference Efficiency.}
We measure end-to-end action-chunk inference
latency on an NVIDIA A100 GPU using the
standard $\pi_{0.5}$ inference implementation,
without method-specific acceleration.
As shown in \Cref{tab:inference_latency},
\genvla requires 527.0 ms per chunk, introducing
14.4\% overhead over static LoRA (460.7 ms),
while remaining 31.3\% lower median latency than PhaseLoRA.
\genvla generates adapters once per replanning
step and reuses them throughout action prediction.
Its Transformer-based generator supports
parallel GPU computation, whereas PhaseLoRA's
recurrent GRU routing may contribute to
additional sequential overhead.
MorphAct~\citep{liu2026morphact} employs two
coupled generators with sequential dependencies
through the adapted VLM, which may also
increase inference latency.
For consistency, we evaluate MorphAct using
the default $\pi_{0.5}$ code without specialized
acceleration.
Static LoRA is evaluated without weight merging;
although its fixed updates can be merged into
the backbone to reduce latency, \genvla's
context-dependent updates cannot be permanently
merged into a single fixed model.

\begin{table}[t]
\centering
\caption{
\textbf{Inference latency on $\pi_{0.5}$.}
Median and 90th-percentile (P90) latency per
action chunk (ms) on an NVIDIA A100 GPU.
All methods use the standard inference path
without specialized acceleration.
LoRA, PhaseLoRA, and \genvla use rank $r=4$.
Static LoRA adapters are not merged.
}
\label{tab:inference_latency}

\footnotesize
\setlength{\tabcolsep}{6pt}
\renewcommand{\arraystretch}{1.15}

\begin{tabular}{lcc}
\toprule
\textbf{Method}
& \textbf{Median (ms) $\downarrow$}
& \textbf{P90 (ms) $\downarrow$} \\
\midrule

Base $\pi_{0.5}$
& 450.8 & 459.3 \\

Static LoRA~\citep{hu2021lora}
& 460.7 & 468.9 \\

PhaseLoRA~\citep{guo2026phaselora}
& 767.4 & 776.6 \\

MorphAct~\citep{liu2026morphact}
& 755.8 & 764.1 \\

\midrule

\textbf{GenVLA (Ours)}
& 527.0 & 536.2 \\

\bottomrule
\end{tabular}
\end{table}

\section{Effective Rank of Generated Updates}
\label{supp:erank}

\subsection{Definition and protocol.}
Let $\Delta W_t$ denote the update to one adapted weight matrix at
sampled frame $t$. All methods in this comparison use $r=4$, so
$\operatorname{rank}(\Delta W_t)\leq 4$. For \genvla{}, main-paper
Eq.~(4) gives $\Delta W_t=B_tA_t$ with $\alpha=r$.
For a nonzero matrix $M$ with positive singular values
$\sigma_1,\ldots,\sigma_q$, the effective rank~\cite{roy2007effective} is
\begin{equation}
  \operatorname{erank}(M)
  = \exp\!\left(-\sum_{i=1}^{q}p_i\log p_i\right),
  \qquad
  p_i=\frac{\sigma_i}{\sum_{j=1}^{q}\sigma_j}.
  \label{eq:supp_erank}
\end{equation}
It ranges from $1$ to $q$ and attains $q$ exactly when all positive
singular values are equal.

The \emph{one-frame} column reports the values in main-paper
Fig.~5(c). For static LoRA and \genvla{}, we average the effective
ranks of individual nonzero updates within each matrix and then
across matrices. For PhaseLoRA, each static VLM matrix contributes
one estimate and each action-expert matrix contributes 24, with
equal weight per estimate.

For \emph{all frames}, we concatenate the updates to each adapted
matrix across $T$ sampled frames,
\begin{equation}
  \mathcal{W}_T =
  \big[\Delta W_1,\ldots,\Delta W_T\big]
  \in \mathbb{R}^{d_{\mathrm{out}}\times (T d_{\mathrm{in}})}.
  \label{eq:supp_concat}
\end{equation}
We compute $\operatorname{erank}(\mathcal{W}_T)$ for each matrix
with a nonzero concatenated update and give each such matrix equal
weight in the reported mean.

For the cross-frame comparison, we re-evaluate PhaseLoRA on the
same 240 LIBERO frames used for \genvla{} in main-paper Fig.~5(c).
The sample contains three frames from each of two training
demonstrations per task across all 40 tasks. This statistic pools
variation within demonstrations and across tasks.

\begin{table}[t]
  \centering
  \footnotesize
  \setlength{\tabcolsep}{3.5pt}
  \caption{\textbf{Effective rank of rank-4 updates on LIBERO.}
    The all-frames column uses $T=240$ frames. The bound applies
    to the algebraic rank of $\mathcal{W}_T$ for each adapted matrix.}
  \label{tab:supp_erank}
  \begin{tabular}{@{}lcccc@{}}
    \toprule
    Method & Update & Bound & One frame & All frames \\
    \midrule
    Static LoRA & $BA$     & $r=4$                       & 2.1 & 2.1   \\
    PhaseLoRA   & $B_tA$   & $4r=16$                     & 2.0 & 2.4   \\
    \genvla{}   & $B_tA_t$ & $\min(d_{\mathrm{out}},Tr)$ & 3.7 & 119.6 \\
    \bottomrule
  \end{tabular}
\end{table}

\subsection{Results and analysis.}
For static LoRA, concatenating $T$ copies of the same update
multiplies each singular value by $\sqrt{T}$. The normalized
singular values therefore stay the same, giving an effective rank
of 2.1 in both columns of \cref{tab:supp_erank}.
In each PhaseLoRA action-expert layer~\cite{guo2026phaselora},
$B_t$ is a linear combination of four learned left factors and
$A$ is shared across frames. The columns of the concatenated
updates thus lie in a subspace of dimension at most $4r$.
The VLM-side updates are static.

For \genvla{}, the mean single-frame effective rank is 3.7 out of 4.
On the same 240 frames, the cross-frame effective rank is 119.6
for \genvla{} and 2.4 for PhaseLoRA. Main-paper Fig.~5(c) displays
the \genvla{} value rounded to 120. Since each frame contributes
at most $r$ output directions, the cross-frame rank is bounded by
$\min(d_{\mathrm{out}},Tr)$, which equals 256 for key/value
projections with $d_{\mathrm{out}}=256$. Main-paper Fig.~5(b)
shows that the cross-frame effective rank of \genvla{} grows during
training under the flow-matching objective in main-paper Eq.~(6),
which has no explicit diversity term.

Even weak singular directions contribute to effective rank.
To test whether the policy benefits from matching the adapter to
the current context, we replace adapters at inference.
Main-paper Table~8 reports 93.8\% LIBERO success for dynamic
\genvla{} under the standard evaluation. With the evaluation code
and episode partitioning used for the interventions, the dynamic
control achieves 89.8\%. Under this same protocol, success falls
to 9.8\% when the initial adapter is held fixed throughout each
episode and to 13.8\% when adapters are shuffled.

The adapter-swap targets in main-paper Fig.~7(b) and Table~9 come
from LIBERO demonstrations used for policy training. For
cross-demonstration swaps, the target demonstrations are excluded
from the bank of replacement adapters. Table~9 also evaluates
replacements from other time points within the same demonstration.
Both analyses show higher mean flow-matching loss with a replacement
adapter than with the target's own adapter.

\section{Real-Robot Experiments}
\label{supp:real_robot}

\subsection{Setup and protocol.}
\begin{figure*}[t]
  \centering
  \includegraphics[width=0.9\textwidth]{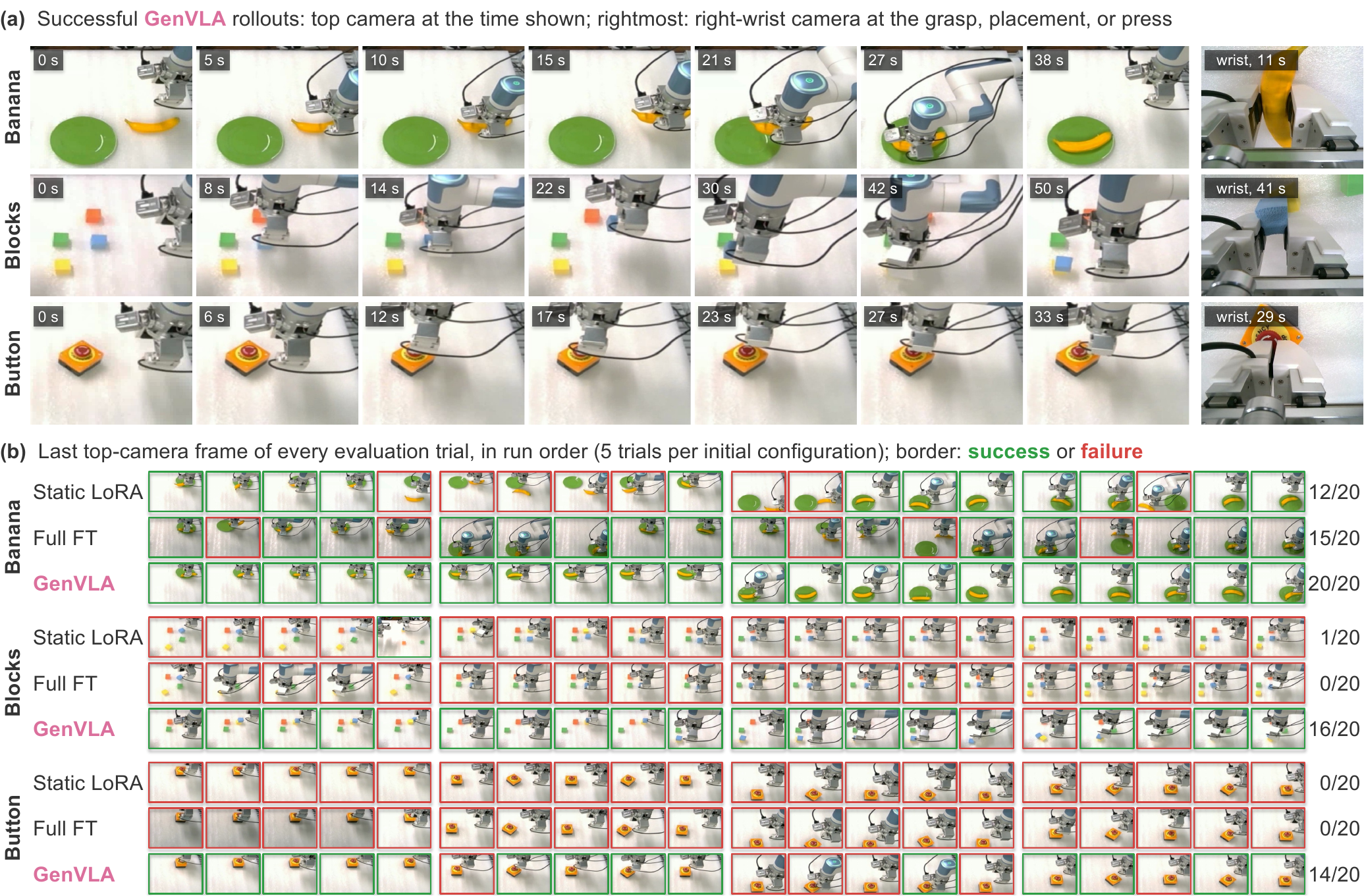}
  \caption{\textbf{Real-robot trials.}
  (a)~One successful \genvla rollout per task.
  (b)~Last top-camera frame of all 180 evaluation trials (3 methods $\times$ 3 tasks $\times$ 20), with the success count of each row.
  Failed baseline trials on block stacking and button pressing typically end with the arm near the target.}
  \label{fig:supp_rr_rollouts}
\end{figure*}

Table~\ref{tab:supp_rr_setup} summarises the platform and tasks, and Fig.~\ref{fig:supp_rr_rollouts} shows example rollouts and the last frame of every trial.
Each method fine-tunes the same $\pi_{0.5}$ checkpoint~\cite{black2025pi05} on the same demonstrations into one multi-task policy for all three tasks. In the real-robot experiments, static LoRA and \genvla each train 8.71\% of their total parameters. Full fine-tuning updates all policy parameters. All methods are evaluated using the same task definitions and initial configurations.
We further analyse the commanded actions and end-effector poses to characterize their execution behaviours.

\begin{table}[t]
  \centering
  \small
  \setlength{\tabcolsep}{4pt}
  \caption{Real-robot setup and tasks, shared by all methods. Quotes give the language prompt.}
  \label{tab:supp_rr_setup}
  \begin{tabular}{@{}l >{\raggedright\arraybackslash}p{0.775\columnwidth}@{}}
    \toprule
    Robot     & Dual-arm Dobot Nova~2 with ViTai GF225 parallel grippers; the policy controls the right arm \\
    Cameras   & Three RealSense RGB cameras (top, both wrists), $640{\times}480$, resized to $224{\times}224$ \\
    Training  & 150 demonstrations of the three tasks at 10\,Hz; 50k steps; chunk $H{=}20$; \genvla history $K{=}3$, $s{=}5$ \\
    Execution & Real-time chunking~\cite{black2026rtc}: a new 20-step chunk every 7 steps, 6-step delay\\
    Trials    & 20 per method and task, 5 in each of 4 initial configurations; at most 300 steps (100\,s); success judged by the operator \\
    \midrule
    Banana    & ``place the banana into the plate'': plate in one of 4 table quadrants, 5 banana placements around it. Success: banana released inside the plate \\
    Blocks    & ``place the cube'': four coloured cubes in 4 layouts. Success: blue cube resting on the yellow cube \\
    Button    & ``press the button'': emergency-stop button at 4 positions, 5 orientations each ($45^\circ$ steps). Success: button pressed \\
    \bottomrule
  \end{tabular}
\end{table}

\subsection{Results and analysis.}
\begin{figure*}[h]
  \centering
  \includegraphics[width=0.9\textwidth]{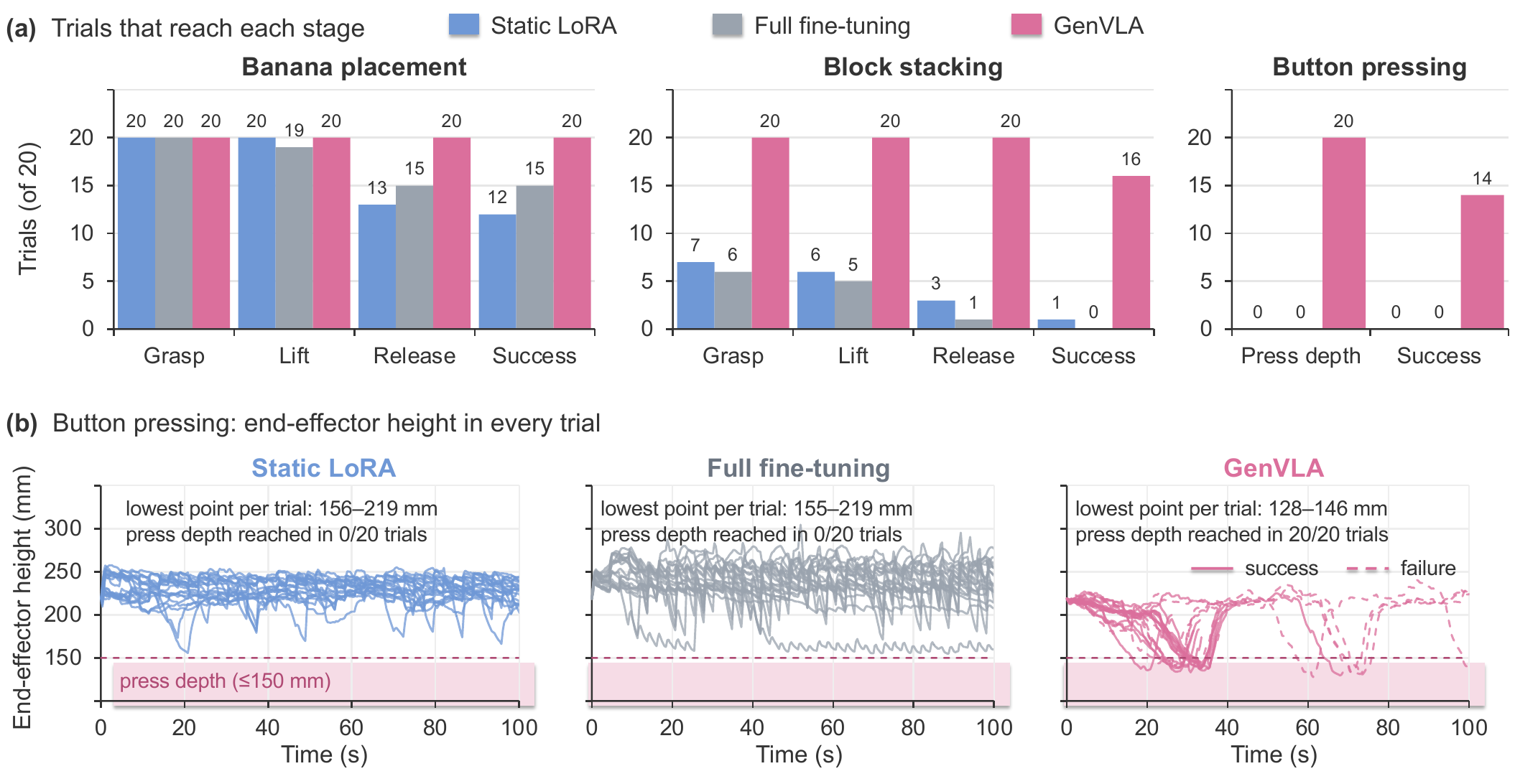}
  \caption{\textbf{Stage-wise analysis from the logs} (20 trials per method and task).
  (a)~Trials that reach each stage.
  \emph{Grasp}: the gripper is commanded closed.
  \emph{Lift}: the end effector then rises by at least 50\,mm before the gripper reopens.
  \emph{Release}: the gripper reopens after a lift.
  \emph{Press depth}: the end effector descends to 150\,mm or lower.
  (b)~End-effector height (robot base frame) in every button trial; dashed: \genvla failures.}
  \label{fig:supp_rr_analysis}
\end{figure*}

Both baselines handle banana placement (12 and 15 successes), where the banana's position reveals the task stage.
On the other two tasks, however, they rarely reach the decisive execution stages (Fig.~\ref{fig:supp_rr_analysis}).
On block stacking, static LoRA succeeds in only 1/20 trials and full fine-tuning in none.
Static LoRA and full fine-tuning close the gripper in 7 and 6 trials and release after a lift in 3 and 1, while \genvla completes both stages in all 20.
On button pressing, no baseline trial descends to pressing depth (lowest points 155--219\,mm, against 128--146\,mm for \genvla); the arm hovers above the button and closes the gripper up to 11 times per trial.
Unsuccessful baseline trials on these two tasks typically run to the 100\,s limit.

We attribute these failures to stage ambiguity.
Above the button, the scene looks almost the same before the press and after the retract, and a missed grasp leaves the cubes as they were; the current frame therefore does not tell an observation-only policy whether to descend or retract, consistent with the hovering and repeated gripper commands.
\genvla instead generates its adapters from the last $K$ executed state--action pairs and vlm features, which record whether the descent or grasp has already happened.
Its own failures are mostly near misses, where the released cube does not stay on the yellow one (four trials) or the gripper reaches the cap without depressing it (six trials).

As prior work also observes~\cite{mandlekar2021matters,kressgazit2024robot,chen2026rmbench}, human involvement introduces variance into real-robot evaluations.
Our demonstrations are collected through teleoperation, introducing variations in approach paths, speeds, and pauses, while manual object resets and success judgments may introduce additional variability.
Nevertheless, all methods learn from the same demonstrations and follow the same evaluation protocol.
The substantial performance gaps on block stacking and button pressing suggest that execution-conditioned parameter generation helps resolve visually ambiguous task stages and supports more reliable multi-stage manipulation.

\clearpage
\appendix


\end{document}